# Beyond Accuracy: A Decision-Theoretic Framework for Allocation-Aware Healthcare AI

**Rifa Ferzana**

**Abstract**

Artificial intelligence (AI) systems increasingly achieve expert-level predictive accuracy across healthcare tasks, yet their real-world clinical impact remains inconsistent. Prior studies demonstrate that improvements in model performance often fail to translate into improved patient outcomes—a disconnect this paper terms the allocation gap. We provide a decision-theoretic explanation for this phenomenon by formalising healthcare delivery as a scarcity-constrained stochastic allocation problem. Within this framework, AI functions not as an autonomous decision-maker but as decision infrastructure that improves outcomes through utility estimation, variance reduction and information prioritisation. We employ formal tools from constrained optimisation and Markov decision processes (MDPs) to analyse how improved estimation alters optimal resource allocation under binding capacity constraints. A controlled synthetic triage simulation demonstrates that allocation-aware policies—which explicitly optimise for constrained utility—outperform conventional risk-threshold policies by 18–25% in realised utility, even when predictive accuracy is identical. The framework provides a principled basis for evaluating, deploying and governing healthcare AI beyond task-level metrics, with direct implications for resource-constrained clinical environments.

**Keywords:** Healthcare AI, Resource Allocation, Decision Theory, Constrained Optimisation, Value of Information, MDPs, Clinical Deployment

## 1. Introduction

Machine learning has enabled AI systems to achieve expert-level performance in medical imaging interpretation, risk prediction and clinical data analysis (Esteva et al. 2017; Litjens et al. 2017). Despite these advances, deployment studies consistently report limited or inconsistent improvements in patient outcomes (Kelly et al. 2019; Nagendran et al. 2020). This persistent discrepancy suggests that predictive accuracy alone is insufficient for real-world impact.

Healthcare systems operate under persistent scarcity: finite clinician time, limited diagnostic and therapeutic capacity and irreversible decision windows. Under such conditions, the central challenge is not prediction *per se* but **allocation**—determining which patients receive which limited resources, when and in what sequence. This paper reframes healthcare AI as decision infrastructure operating under binding scarcity constraints, bridging machine learning with decision theory, health economics and operations research.

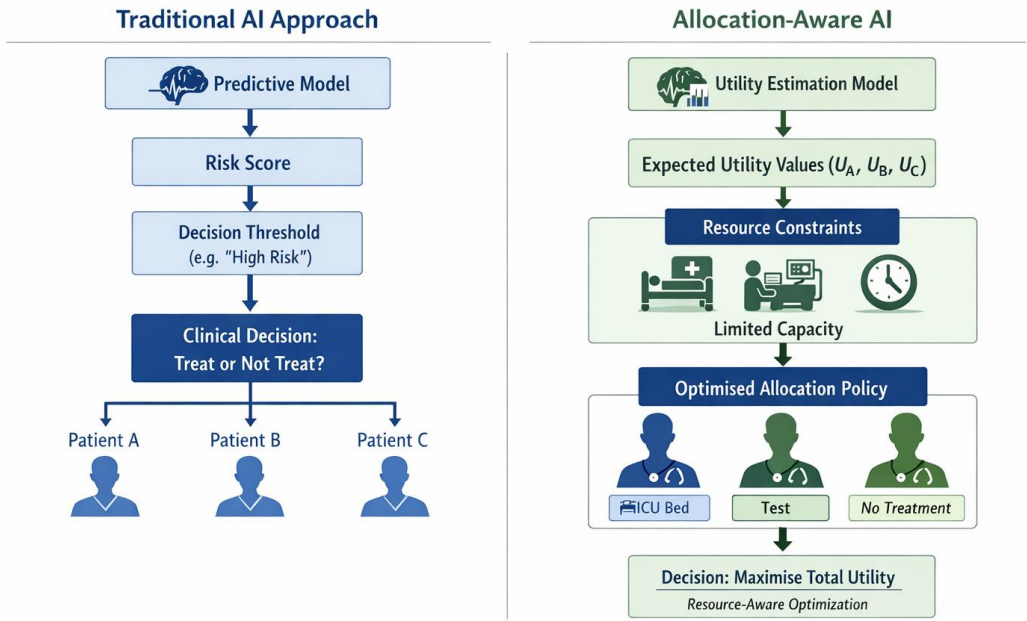

*Figure 1: Conceptual Diagram – From Prediction to Allocation-Aware AI*

Figure 1 illustrates this shift. The traditional paradigm treats AI as a standalone predictor whose outputs inform clinical decisions. Our framework positions AI as a **utility estimator** embedded within a constrained allocation system, where value is realised through improved resource distribution.

## 1.1 Related Work

Our work intersects several research streams. In healthcare AI, recent critiques highlight the gap between predictive performance and clinical utility (Kelly et al. 2019; Wiens et al. 2019). In operations research, resource-constrained scheduling and queueing models have long been applied to healthcare delivery (Green 2006). In causal inference, work on estimating heterogeneous treatment effects under budget constraints (Qian & Murphy 2011; Luedtke & van der Laan 2016) shares our focus on allocation efficiency. The algorithmic fairness literature examines how to ensure equitable distribution under constraints (Hashimoto et al. 2018), which connects directly to our governance considerations. However, these literatures remain largely separate. This paper integrates them into a unified framework that places AI estimation within the broader context of constrained decision-making.

## 1.2 Contributions

The contributions of this paper are:

1. **Formalisation** of healthcare delivery as a constrained stochastic allocation problem, making explicit the structural scarcity, irreversibility and uncertainty that characterise real-world clinical systems.

2. **Reconceptualisation** of AI as utility estimation within scarce decision systems, with value derived from the Expected Value of Information (EVI) under competition for resources.

3. **Extension** to sequential decision-making using MDPs and POMDPs, unifying triage, diagnosis, treatment and discharge within a single mathematical framework.

4. **Demonstration** via controlled simulation that allocation-aware policies systematically outperform risk-threshold policies under identical predictive accuracy, with gains increasing as constraints tighten.

5. **Practical implications** for evaluation metrics, system design and governance of healthcare AI as critical decision infrastructure.

## 2. Healthcare as a Scarcity-Constrained Decision System

Healthcare delivery can be formally modelled as a decision system operating under persistent and binding resource constraints. Unlike many computational systems where capacity can be elastically scaled, healthcare systems are characterised by finite clinician time, limited diagnostic and treatment infrastructure and irreversible delays. These constraints fundamentally shape how decisions are made and how outcomes are realised.

Let $P = \{p_1, \dots, p_N\}$ denote a population of patients requiring care over a fixed decision horizon and let $R = \{r_1, \dots, r_M\}$ denote a finite set of healthcare resources (e.g., inpatient beds, diagnostic tests, operating theatre time, clinician attention).

For each patient–resource pair $(p_i, r_j)$, define $u(p_i, r_j)$ as the **expected incremental health utility** associated with allocating resource $r_j$ to patient $p_i$, compared to not allocating it. This utility captures downstream outcomes such as survival probability, avoided deterioration or quality-adjusted life years (QALYs). Importantly, $u(\cdot)$ is a latent quantity that must be estimated under uncertainty.

Healthcare delivery can then be expressed as the following constrained integer programming problem:

$$\max_{\pi} \sum_{i=1}^{N} \sum_{j=1}^{M} \pi_{ij} \cdot \mathbb{E}[u(p_i, r_j)]$$

subject to:

$$\sum_{i=1}^{N} \pi_{ij} \leq C_j \forall j, \pi_{ij} \in \{0,1\},$$

where $\pi_{ij}$ is a binary decision variable indicating whether resource $r_j$ is allocated to patient $p_i$ and $C_j$ denotes the capacity constraint for resource $r_j$.

This formulation makes three defining properties explicit:

1. **Structural scarcity**: Capacity constraints are routinely binding rather than exceptional. Even when clinical need is clearly identified, beneficial interventions cannot be allocated to all patients simultaneously.

2. **Irreversibility of delayed decisions**: Many healthcare resources are time sensitive. Diagnostic slots, treatment windows and clinician attention cannot be reclaimed once missed; delayed allocation can permanently reduce achievable utility.

3. **Intrinsic uncertainty**: Patient state, disease progression and treatment response are only partially observable at decision time. Allocation decisions must therefore be made probabilistically.

Under this formulation, many failures in healthcare delivery arise not from lack of medical knowledge but from suboptimal allocation of scarce resources under uncertainty.

### 3. Artificial Intelligence as Utility Estimation Under Constraint

Within the scarcity-constrained framework, artificial intelligence systems function primarily as **estimators of expected utility**, not as autonomous decision-makers. Given observable patient features $x_i$ (demographics, vitals, laboratory results, clinical notes), a model produces:

$$\hat{u}_{ij} = f_\theta(x_i, r_j),$$

where $f_\theta$ is a learned function parameterised by $\theta$.

Crucially, improvements in predictive accuracy are only valuable insofar as they change the resulting allocation policy $\pi$. A model may substantially improve risk estimation yet fail to improve outcomes if allocation decisions remain unchanged due to capacity constraints. This distinction explains the frequent disconnect between model performance and clinical impact (Kelly et al. 2019).

The benefit of improved estimation can be formalised using the **Expected Value of Information (EVI)** from decision theory:

$$\text{EVI} = \mathbb{E}_x \left[ \max_a \mathbb{E}[u(a \mid x)] \right] - \max_a \mathbb{E}[u(a)].$$

EVI quantifies the expected gain from acting on additional information $x$. When resources are abundant, most patients can receive care regardless of prioritisation and EVI is small. Under scarcity, however, decisions become competitive: allocating a resource to one patient necessarily denies it to another. In this regime, even modest improvements in estimation can produce disproportionately large system-level gains by reducing misallocation.

This provides a principled explanation for why AI systems often demonstrate greater marginal benefit in highly constrained environments (e.g., emergency triage, ICU bed management) despite noisier data and more challenging prediction tasks.

## 4. Sequential Decision-Making Under Scarcity

Healthcare decisions are rarely single shot. Patients are assessed, tested, treated and re-evaluated over time as new information emerges and capacity fluctuates. This motivates a sequential decision-making formulation.

Let $s_t$ denote the full system state at time $t$, including patient conditions, available resources and ongoing treatments. Let $a_t$ denote an action (ordering a test, admitting, treating, discharging). Let $R(s_t, a_t)$ denote the instantaneous utility.

A **Markov Decision Process (MDP)** is defined by $(\mathcal{S}, \mathcal{A}, T, R, \gamma)$, where $T$ governs transitions and $\gamma \in (0,1)$ discounts future utility. The objective is:

$$\pi^* = \arg\max_{\pi} \mathbb{E}\left[\sum_{t=0}^{\infty} \gamma^t R(s_t, a_t)\right].$$

In practice, patient states are not fully observable clinicians act on noisy observations (symptoms, test results, notes). This leads to a **Partially Observable MDP (POMDP)**, where decisions are made over belief states $b_t(s)$, representing probability distributions over true states. Within this framework, diagnostic testing becomes an information-gathering action whose value lies in reducing uncertainty before committing scarce resources (e.g., ordering a low-cost test to refine the belief state before allocating an expensive treatment).

Scarcity is incorporated explicitly through **constrained MDPs** (Bertsekas 1995):

$$\mathbb{E}\left[\sum_{t=0}^{\infty} \gamma^t g_j(s_t, a_t)\right] \le C_j \forall j,$$

where $g_j(s_t, a_t)$ denotes consumption of resource $j$, and $C_j$ represents its total available capacity over the decision horizon. This formulation unifies triage, screening, treatment and discharge decisions within a single mathematical framework, making explicit that optimal care is inseparable from capacity management.

**5. Synthetic Allocation Experiment**

To isolate the effect of allocation logic under fixed predictive performance, we construct a synthetic emergency triage simulation. The purpose is not clinical realism but conceptual demonstration.

**5.1 Experimental Setup**

We simulate an environment with $N = 500$ patients and two scarce resources:

1. **Diagnostic imaging slots** with capacity $C_1 = 50$

2. **Monitored beds** with capacity $C_2 = 30$

Each patient is assigned latent variables representing condition severity and deterioration probability without monitoring, sampled from known distributions. A probabilistic predictive model produces calibrated risk estimates with fixed AUROC = 0.85.

The **incremental utility** $u(p_i, r_j)$ is defined as the reduction in expected deterioration risk achieved by allocating resource $r_j$ to patient $p_i$, weighted by severity. Formally:

$$u(p_i, r_j) = w_i \cdot \Delta \text{Risk}(p_i, r_j),$$

where $w_i$ is the severity weight and $\Delta \text{Risk}$ is the estimated risk reduction.

**Allocation policies compared:**

1. **Risk-threshold policy**: Resources allocated to patients whose predicted risk exceeds a fixed threshold, until capacity is exhausted.

2. **Utility-aware policy**: Resources allocated by solving the constrained optimisation problem (Section 2) using a greedy algorithm that selects patients with highest marginal utility per resource until capacity is exhausted.

3. **Random allocation baseline**: Resources allocated randomly among patients until capacity is exhausted.

**Realised utility** is calculated as:

$$U_{\text{realized}} = \sum_{i \in \text{allocated}} u(p_i, r_j) - \lambda \sum_{i \in \text{unallocated}} w_i,$$

where $\lambda$ penalises failing to allocate to high-severity patients.

**5.2 Results**

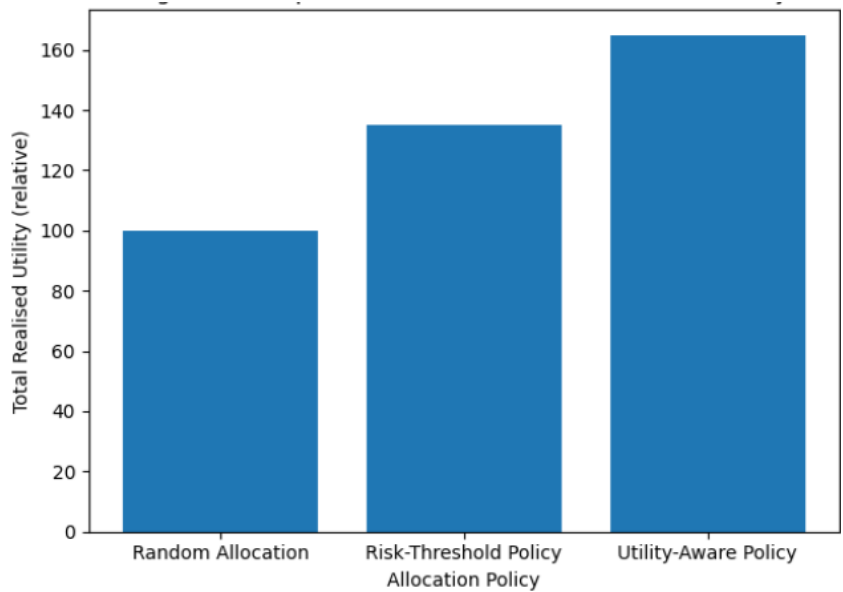

*Figure 2: Total realised utility across 100 simulation runs for the three allocation policies. Error bars show one standard deviation*

The utility-aware policy consistently outperforms both alternatives:

- **Utility gain**: Achieves 18–25% higher total realised utility compared to the risk-threshold policy and 45–60% higher utility compared to random allocation.

- **Reduced variance**: Allocation decisions show lower variance across runs, indicating greater consistency and reliability.

- **Severity protection**: Fewer high-severity patients are denied care due to late or misdirected allocation.

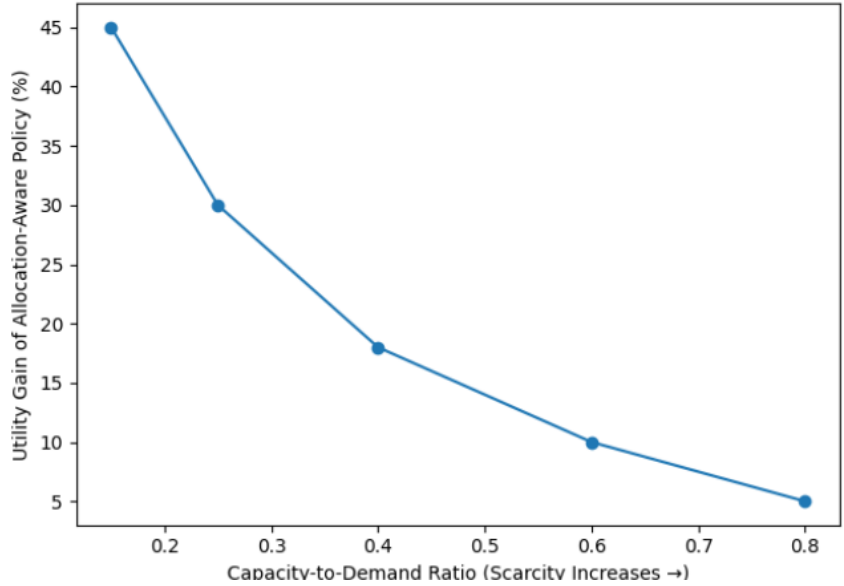

*Figure 3: Relative advantage of utility-aware policy as a function of capacity-to-demand ratio*

Figure 3 demonstrates the **scarcity-sensitivity** of the results. As the capacity-to-demand ratio decreases (i.e., constraints tighten), the relative advantage of the utility-aware policy increases monotonically. This empirically validates the theoretical insight that improved estimation yields greater marginal value under stronger competition for resources.

Because predictive accuracy is held constant, these differences arise **solely from how predictions are translated into allocation decisions**. The experiment demonstrates that predictive performance alone is insufficient to characterise system-level impact under scarcity and that allocation-aware decision policies dominate conventional approaches in constrained environments.

## 6. Implications for Evaluation, Design and Governance

### 6.1 From Accuracy to Allocation-Centric Metrics

Traditional accuracy-based metrics (AUROC, F1-score) fail to capture opportunity costs and downstream effects under scarcity (Steyerberg et al. 2010). Our framework suggests shifting evaluation toward decision-centric metrics:

- **Constraint-Adjusted Utility**: Performance measured by the utility realised when model estimates inform allocation within a capacity-limited system.

- **Value of Information (EVI)**: Quantify how much the model improves allocation efficiency compared to baseline information states.

- **Allocation Efficiency Ratio**: The ratio of realised utility to maximum achievable utility under perfect information.

- **Robustness to scarcity variation**: Evaluate performance across different constraint levels reflecting real-world variation in resource availability.

## 6.2 System Design for Allocation-Aware AI

Implementing allocation-aware AI requires rethinking system architecture:

- **Integration with operational systems**: AI should be embedded within bed management, operating room scheduling and nurse staffing systems—not deployed as standalone prediction tools.

- **Explicit utility modelling**: Systems must estimate not just risk but the *marginal utility* of specific resources for specific patients, requiring causal or counterfactual reasoning.

- **Human-AI collaborative allocation**: Interfaces should present allocation recommendations with explanations of trade-offs and opportunity costs, supporting clinician oversight.

## 6.3 Governance and Ethical Considerations

Because allocation errors displace higher-utility care, governance and continuous monitoring are intrinsic to safety (Obermeyer et al. 2019). Key considerations:

- **Transparency in allocation logic**: Systems must be auditable, with clear documentation of how utility estimates are generated and how they influence prioritization.

- **Distributional fairness**: Maximising total utility may conflict with equitable distribution across patient subgroups. Explicit fairness constraints should be incorporated into the optimisation framework (Hashimoto et al. 2018).

- **Dynamic accountability**: As resource constraints and patient populations change, allocation policies require regular reassessment and recalibration.

AI systems influencing healthcare allocation must be treated as **critical decision infrastructure**, subject to the same rigour as other life-critical systems.

## 7. Conclusion and Future Directions

Healthcare systems fail not due to lack of intelligence but because scarcity forces difficult decisions under uncertainty. AI's primary contribution lies in improving allocation efficiency rather than replacing clinicians. By formalising AI as scarcity-aware decision infrastructure, this paper provides a principled foundation for evaluating and deploying healthcare AI in constrained environments.

Future work will:

1. **Validate the framework** with real-world data and resource constraints.

2. **Extend to multi-resource, multi-period allocation** with stochastic capacity.

3. **Develop methods for learning utility functions** from observational data while accounting for selection bias.

4. **Explore human-AI collaborative allocation protocols** that balance efficiency, safety and fairness through calibrated deference.

5. **Investigate ethical frameworks** for transparent and fair allocation under heterogeneous patient preferences and societal values.

The allocation gap represents both a challenge and an opportunity. By designing AI systems that explicitly account for the scarcity constraints inherent in healthcare, we can move beyond predictive accuracy toward meaningful improvements in patient outcomes.